\documentclass{article}

\usepackage{arxiv}

\usepackage{graphicx}
\usepackage[utf8]{inputenc} 
\usepackage[T1]{fontenc}    
\usepackage[colorlinks,allcolors=blue]{hyperref}       
\usepackage{url}            
\usepackage{booktabs}       
\usepackage{amsfonts}       
\usepackage{nicefrac}       
\usepackage{microtype}      
\usepackage{lipsum}		
\usepackage{graphicx}
\usepackage{natbib}
\usepackage{doi}
\usepackage{tabularx,booktabs}
\usepackage{algorithm}
\usepackage{algpseudocode} 
\usepackage{multirow}
\usepackage{xcolor}
\usepackage{multicol}
\usepackage{subcaption}
\usepackage{tikz} 
\usepackage{amsmath}
\usepackage{textcomp}
\usepackage{libertine}
\usepackage{comment}
\usepackage{amssymb}

\algnewcommand\algorithmicforeach{\textbf{for each}}
\algnewcommand\algorithmicdoparallel{\textbf{do in parallel}}
\algdef{S}[FOR]{ForEach}[1]{\algorithmicforeach\ #1\ \algorithmicdo}
\algdef{S}[FOR]{ForEachParallel}[1]{\algorithmicforeach\ #1\ \algorithmicdoparallel}
\algnewcommand{\sIf}[2]{
  \State \algorithmicif\ #1\ \algorithmicthen\ #2}
\newcommand{\assign}{\ensuremath{\leftarrow}}
\newcommand{\true}{\textsc{true}}
\newcommand{\false}{\textsc{false}}

\graphicspath{graphics}
\usepackage{xspace}
\newcommand{\Eoffset}{\ensuremath{E_{\text{offset}}}}
\newcommand{\MaxIterations}{\ensuremath{I_{\text{max}}}}
\newcommand{\offsetIncreaseRate}{\beta}
\newcommand{\nondom}{\ensuremath{\parallel}}

\title{Multi-objective QUBO Solver: Bi-objective Quadratic Assignment Problem\thanks{Please cite: \protect\doi{https://doi.org/10.1145/3512290.3528698}}}

\author{ \href{https://orcid.org/0000-0003-0854-4777}{\includegraphics[scale=0.06]{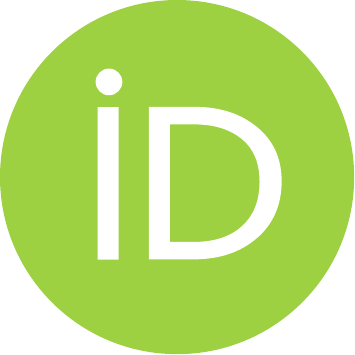}\hspace{1mm}Mayowa Ayodele}\\
	Fujitsu Research of Europe Ltd.\\
	The Urban Building, 3-9 Albert Street\\
	Slough, United Kingdom,  SL1 2BE\\
	\texttt{mayowa.ayodele@fujitsu.com} \\
	\And
	\href{https://orcid.org/0000-0003-1236-3143}{\includegraphics[scale=0.06]{orcid.pdf}\hspace{1mm}Richard Allmendinger} \\
	The University of Manchester\\
	Manchester\\
	United Kingdom \\
	\texttt{richard.allmendinger@manchester.ac.uk} \\
		\And
	\href{https://orcid.org/0000-0001-9974-1295}{\includegraphics[scale=0.06]{orcid.pdf}\hspace{1mm}Manuel López-Ibáñez} \\
	The University of Manchester\\
	Manchester\\
	United Kingdom \\
	\texttt{manuel.lopez-ibanez@manchester.ac.uk} \\
		\And
	\href{https://orcid.org/0000-0002-5777-7756}{\includegraphics[scale=0.06]{orcid.pdf}\hspace{1mm}Matthieu Parizy} \\
	Fujitsu Laboratories Ltd.\\
	Kawasaki\\
	Japan \\
	\texttt{parizy.matthieu@fujitsu.com} \\
}

\renewcommand{\shorttitle}{Multi-objective QUBO Solver: BQAP}

\begin{document}
\maketitle

\begin{abstract}
  Quantum and quantum-inspired optimisation algorithms are designed to solve problems represented in binary, quadratic and unconstrained form. Combinatorial optimisation problems are therefore often formulated as Quadratic Unconstrained Binary Optimisation Problems (QUBO) to solve them with these algorithms. Moreover, these QUBO solvers are often implemented using specialised hardware to achieve enormous speedups, e.g. Fujitsu's Digital Annealer (DA) and D-Wave's Quantum Annealer. However, these are single-objective solvers, while many real-world problems feature multiple conflicting objectives. Thus, a common practice when using these QUBO solvers is to scalarise such multi-objective problems into a sequence of single-objective problems. Due to design trade-offs of these solvers, formulating each scalarisation may require more time than finding a local optimum. We present the first attempt to extend the algorithm supporting a commercial QUBO solver as a multi-objective solver that is not based on scalarisation. The proposed multi-objective DA algorithm is validated on the bi-objective Quadratic Assignment Problem. We observe that algorithm performance significantly depends on the archiving strategy adopted, and that combining DA with non-scalarisation methods to optimise multiple objectives outperforms the current scalarised version of the DA in terms of final solution quality.

\end{abstract}




\section{Introduction}
There has been significant research interest in quantum-inspired and quantum optimisation algorithms in recent years. Such algorithms are able to exploit the use of specialised hardware to solve optimisation problems much quicker than classical algorithms implemented on general purpose machines \citep{albash2018demonstration, aramon2019physics}.

Quantum-inspired as well as quantum algorithms are designed for solving problems that are in binary and quadratic form only. Quadratic Unconstrained Binary Optimisation (QUBO), also sometimes referred to as Unconstrained Binary Quadratic Programming (UBQP) or ising model, is one of the most widely used representations when solving combinatorial optimisation problems \citep{VerLew2020penalty}. Knapsack, travelling salesman and graph partitioning problems can be formulated as QUBO problems~\citep{Lucas2014ising}.

Commercial QUBO solvers such as D-wave's Quantum Annealer (QA) and Fujitsu's Digital Annealer (DA) have evolved in the last ten years. Fujitsu's first generation DA, introduced in 2018, was able to optimise QUBOs with up to 1,024 bits. The third generation DA is now able to optimise Binary Quadratic Problems (BQPs) with up to 100,000 bits \citep{da3}. BQPs include QUBO and other binary and quadratic formulations that may have constraints. D-wave One released in 2011 was able to optimise QUBOs with 128 quantum bits (qubits). D-wave's Advantage released in 2020 contains at least 5,000 qubits \citep{dwaveadvantage}. D-Wave's Advantage is quantum based \citep{dwaveadvantage} whereas DA is CMOS hardware based~\citep{aramon2019physics}.

In addition to quantum annealing and digital annealing algorithms used in commercial solvers, other algorithms such as estimation of distribution algorithms, scatter search, simulated annealing (SA), tabu search and genetic algorithm have also been applied to QUBOs \citep{Beasley1998,KocHaoGlo2014bqap}.

Multi-objective optimisation has been a topic of interest for many years in operations research \citep{Wierzbicki1980mcdmta}. There are various approaches of solving multi-objective problems; a priori (e.g scalarisation,  goal programming, $\epsilon$-constraint), a posteriori, and interactive \citep{Coello2000cec,Deb:MOEA}. 
Although a posteriori algorithms, based on Pareto optimality, have been designed to solve QUBOs with multiple objectives~\citep{FujNan2021solving,LieVerPaqHao2015bubqp}, 
a priori methods are used when solving multi-objective problems with the QA or DA since they are single-objective solvers. For example, scalarisation was used to solve a bi-objective listing optimisation problem with the QA \citep{NisTanSugMiy2019item} and to solve the bi-objective problem of routing and wavelength assignment with the DA \citep{cseker2020routing}. The $\epsilon$-constraint approach was used with the QA to solve the portfolio optimisation problem~\citep{phillipson2021portfolio}.

In this study, we propose a new multi-objective DA (MDA) algorithm for Pareto optimisation by redesigning the existing DA algorithm, which is based on local search, to handle multiple objectives and an archive of solutions. The first step to extending single-objective local search to multi-objective optimisation is to update the acceptance criteria \citep{PaqChiStu2004mmo}. Since the DA is similar to SA, we adapt and compare two acceptance criteria previously proposed for multi-objective SA \citep{amine2019multiobjective}. Different methods of updating the archive are also compared. Finally, we compare a scalarisation-based single-objective DA (SB-DA) with MDA. Due to the peculiarities of the DA algorithm, in particular, its extremely fast neighbourhood exploration but slow setup for individual scalarisations, conclusions obtained in previous analyses of multi-objective local search methods may not apply to the MDA.

As a benchmark problem, we consider the bi-objective Quadratic Assignment Problem (QAP). The DA has been shown to present competitive performance on single-objective QAP instances \citep{cseker2020digital,MatTakMiyShi2020digital}. 
In \citep{MatTakMiyShi2020digital}, the DA was able to solve QAPLIB \citep{BurKarRen1997} instances to optimality up to 165,000 times faster than CPLEX.

The rest of this paper is structured as follows. A formal description as well as the QUBO formulation of the bi-objective QAP is presented in Section~\ref{sec:prob}. Details of the proposed algorithm for multi-objective optimisation within DA are presented in Section~\ref{sec:da}. Problem instances, parameter settings as well as performance measure used in this study are described in Section~\ref{sec:exps}. Results are analysed in Section \ref{sec:results}. Finally, conclusions are presented in Section \ref{sec:con}.

\section{Problem Description}
\label{sec:prob}
In this section, the bi-objective problem is formally defined, and the QUBO formulation is presented.

\subsection{Single-Objective Quadratic Assignment Problem}
The Quadratic Assignment Problem (QAP) \citep{KooBec57} can be described as the problem of assigning a set of $n$ facilities to a set of $n$ locations. For each pair of locations, a distance is specified. For each pair of facilities, a flow (or weight) is specified. The aim is to assign each facility to a unique location such that the sum of the products between flows and distances is minimised. 

Formally, an instance of the QAP consists of two $n \times n$ input matrices $H=[h_{ij}]$ and $D=[d_{uv}]$, where $h_{ij}$ is the flow between facilities $i$ and $j$, and $d_{uv}$ is the distance between locations $u$ and $v$. A solution to the QAP is a permutation $\sigma = (\sigma_1, \dotsc ,\sigma_n )$ where $\sigma_i$ represents the location assigned to facility $i$. The cost function is formally defined as follows
\begin{equation}
    \text{minimise}\ f(\sigma) = \sum_{i=1}^{n}\sum_{j=1}^{n}h_{ij} \cdot d_{\sigma_i\sigma_j}\;.
    \label{eq:nqap}
\end{equation}

\subsection{QUBO formulation of the QAP}

 QUBO problems are unconstrained, quadratic and of binary form generally  defined as follows:
\begin{equation}
\label{eq:qubo}
  E(x) = x^TQx + q \;,
\end{equation}
where $Q$ represents a $m \times m$ matrix, $q$ is a constant term, a solution $x=(x_1, \dots,x_m)$, $x_i\in\{0,1\}$, is an $m$-dimensional binary vector, and $E(x)$ is the \emph{energy} (or fitness) of $x$. 

The QAP can be formulated as QUBO using Eq. \eqref{eq:cgqqap} where the cost function $c(x)$ and the constraint function $g(x)$ are presented in Eqs.~\eqref{eq:cqap} and \eqref{eq:gqap}, respectively. The penalty weight is denoted by $\alpha$ and set using the method presented later in Section~\ref{sec:sbda1} (Eq. \eqref{eq:wvermaa}).
\begin{equation}
\label{eq:cgqqap}
\begin{split}
   E(x) = c(x) + \alpha \cdot g(x)
\end{split}
\end{equation}
\begin{equation}
\label{eq:cqap}
 c(x) = \sum_{i=1}^{n}\sum_{j=1}^{n}\sum_{u=1}^{n}\sum_{v=1}^{n}h_{ij}d_{uv}x_{iu}x_{jv}
\end{equation}
\begin{equation}
\label{eq:gqap}
   g(x) = \sum_{i=1}^{n}\left ( 1- \sum_{u=1}^{n} x_{iu}\right )^2 +\sum_{u=1}^{n}\left ( 1- \sum_{i=1}^{n} x_{iu}\right )^2 
 \end{equation}

 In Eq. \eqref{eq:nqap}, the fitness function $f(\sigma)$ is equivalent to cost function $c(x)$ in \eqref{eq:cgqqap}, where $\sigma$ is a solution represented as a permutation and $x$ is the equivalent two-way one-hot (permutation matrix) solution. The two-way one-hot representation is often used to encode permutation problems when transforming them to QUBO \citep{Lucas2014ising, liu2019modeling}. In this representation, each entity with index $i$ (e.g. location $i$ in the QAP) is represented by a substring of $n$ zeros but the $i^\text{th}$ bit is set to 1, thus a binary variable $x_{iu}$ indicates whether facility $u$ is assigned to location $i$ or not. To prevent solutions that cannot be decoded to a valid permutation, $g(x)$ penalises any solution where each row and/or column does not sum to one (Eq.~\ref{eq:gqap}).  The function $g(x)$ is non-negative: either $g(x)=0$ if the solution $x$ can be decoded to a valid permutation or $g(x)>0$ for infeasible solutions. The value of $g(x)$ increases according to the degree of constraint violation.

 An example of a valid permutation and its encoding is
\begin{equation}
\sigma = (3, 1, 2)  \Leftrightarrow  x = \begin{bmatrix}
0 & 0 & 1\\ 
1 & 0 & 0\\ 
0 & 1 & 0
\end{bmatrix}
\label{eq:x2}  
\end{equation}

Typically $x$ is represented in vector format, where $x$ in Eq.~\eqref{eq:x2} is the same as
\begin{equation}
x = (0 ,0, 1 , 1 , 0 , 0 , 0 , 1 , 0)\;, 
\label{eq:x1}
\end{equation}
meaning that QUBO solutions have dimension $m = n^2$.

As an example, consider the following QAP instance:
\begin{align}
H = \begin{bmatrix}
0 & 1 & 2\\ 
1 & 0 & 1\\ 
2 & 1 & 0
\end{bmatrix}, \ 
&\qquad
D = \begin{bmatrix}
0 & 3 & 4\\ 
3 & 0 & 6\\ 
4 & 6 & 0
\end{bmatrix}\enspace,
\end{align}
for which the corresponding QUBO matrix $C$ and the constant term $p_c$ representing the cost function are presented as follows:

\begin{align}
 C = \begin{bmatrix}
 0&  0&  0&  0&  6&  8&  0& 12& 16  \\       
 0&  0&  0&  6&  0& 12& 12&  0& 24  \\       
 0&  0&  0&  8& 12&  0& 16& 24&  0  \\       
 0&  0&  0&  0&  0&  0&  0&  6&  8  \\       
 0&  0&  0&  0&  0&  0&  6&  0& 12  \\       
 0&  0&  0&  0&  0&  0&  8& 12&  0  \\       
 0&  0&  0&  0&  0&  0&  0&  0&  0  \\       
 0&  0&  0&  0&  0&  0&  0&  0&  0  \\       
 0&  0&  0&  0&  0&  0&  0&  0&  0
\end{bmatrix}, \qquad p_c = 0 \nonumber 
\label{eq:qc}
\end{align}

and the QUBO matrix $G$ and the constant term $p_g$ representing the constraint function are presented as follows:
\begin{align}
  G = \begin{bmatrix}
-2&  2&  2&  2&  0&  0&  2&  0&  0\\
 0& -2&  2&  0&  2&  0&  0&  2&  0\\
 0&  0& -2&  0&  0&  2&  0&  0&  2\\
 0&  0&  0& -2&  2&  2&  2&  0&  0\\
 0&  0&  0&  0& -2&  2&  0&  2&  0\\
 0&  0&  0&  0&  0& -2&  0&  0&  2\\
 0&  0&  0&  0&  0&  0& -2&  2&  2\\
 0&  0&  0&  0&  0&  0&  0& -2&  2\\
 0&  0&  0&  0&  0&  0&  0&  0& -2
\end{bmatrix}, \qquad  p_g = 6 \nonumber\\
\end{align}

An example of a valid solution (i.e valid permutation) is
\begin{multline*}
\sigma = [3,1,2],\ x = (0 ,0, 1 , 1 , 0 , 0 , 0 , 1 , 0)  \\
f(\sigma) = 0\cdot0\ +\ 1\cdot4\ +\ 2\cdot6\ + \ 1\cdot4\ +\ 0\cdot0\ + 1\cdot3\ +\ 2\cdot6\ + 1\cdot3\ +\ 0\cdot0\ = 38 \\
c(x) = x^TCx + p_c = 38 + 0 = 38, \qquad
g(x) = x^TGx + p_g = -6 + 6 = 0
\end{multline*}
while an invalid solution would be
\begin{multline*}
\sigma = [3,2,2],\ x = (0 ,0, 1 , 0 , 1 , 0 , 0 , 1 , 0)  \\
f(\sigma) = 0\cdot0\ +\ 1\cdot6\ +\ 2\cdot6\ + \ 1\cdot6\ +\ 0\cdot0\ +\ 1\cdot0\ +\ 2\cdot6\ + 1\cdot0\ +\ 0\cdot0\ = 36\\
c(x) = x^TCx + p_c = 36 + 0 = 36, \qquad
g(x) = x^TGx + p_g = -4 + 6 = 2
\end{multline*}
In both cases $c(x) \equiv f(\sigma)$.


\subsection{Bi-Objective Quadratic Assignment Problem}
The multi-objective QAP has real-world relevance in scenarios like the hospital layout problem, where it may be important to simultaneously minimise the flows of doctors, patients, hospital guests, etc \citep{KnoCor2003emo}. In this study, we consider the bi-objective QAP.

An instance of the bi-objective QAP consists of a $2 \times n \times n$ flow matrix $H=[h_{kij}]$ and a $n \times n$ distance matrix $D=[d_{uv}]$.
The cost function of the bi-objective QAP is defined as
\begin{equation}
\begin{split}
   \text{minimise}\ \vec{f}(\sigma) &= \left (f_1(\sigma), f_2(\sigma)\right )\\
   \text{where}\ f_k(\sigma) &= \sum_{i=1}^{n}\sum_{j=1}^{n}h_{kij} \cdot d_{\sigma_i\sigma_j}, \quad k \in \left \{ 1,2 \right \}\;.
 \end{split}
\end{equation}

The equivalent QUBO formulation of $f_k(\sigma)$  is
%
\begin{equation}
\label{eq:bf}
\begin{split}
 c_k(x) = \sum_{i=1}^{n}\sum_{j=1}^{n}\sum_{u=1}^{n}\sum_{v=1}^{n}h_{kij}d_{uv}x_{iu}x_{jv},\ k \in \left \{ 1,2 \right \}\;.\\
   \end{split}
\end{equation}

When a two-way one-hot solution $x$ is equivalent to a permutation $\sigma$, then  $f_k(\sigma) \equiv c_k(x)$. The constraint function $g(x)$ (Eq.~\ref{eq:wqap}) remains the same as that of the single-objective formulation (Eq.~\ref{eq:gqap}). We explore solving the QUBO formulation of the bi-objective QAP using a single-objective algorithm and a multi-objective algorithm. In the single-objective solution approach, the weighted sum method is used as follows
\begin{equation}
\label{eq:wqap}
\text{minimise}\ E(x) = \gamma \cdot c_1(x) + (1- \gamma) \cdot c_2(x) + \alpha \cdot g(x),\; \gamma \in [0,1]\;.
\end{equation}

The formulation for the multi-objective (Pareto optimisation) solution approach is.
\begin{equation}
\label{eq:mqap}
\text{minimise }\vec{E}(x) = (c_1(x) + \alpha_1 \cdot g(x),  c_2(x) + \alpha_2 \cdot g(x))
\end{equation}

\section{Digital Annealer}
\label{sec:da}

In this section, we present the SB-DA as well as the proposed MDA algorithm. The algorithm that supports the first generation DA~\citep{aramon2019physics} is used in this study. In particular, we use a CPU implementation of Alg.~\ref{alg:sbda} and Alg.~\ref{alg:moda0} to solve the bi-objective QAP when formulated as a single-objective problem and multi-objective problem, respectively.

\subsection{Digital Annealer Algorithm}

\begin{algorithm}[b]
\caption{DA Algorithm}\label{alg:da1}
\begin{algorithmic}[1]
\Require $Q$, $\MaxIterations$, $\delta_0$, $\delta_f$, $\xi$, $\offsetIncreaseRate$
\State $x \assign$ a random binary solution of length $m$
\State $\delta  \assign  \delta_0$
\For {iteration = 1 to $\MaxIterations$ }
\sIf{$\delta > \delta_f$}{$\delta \assign  \delta \cdot\left ( 1-\xi\right )$}\label{alg:dacool}
\State calculate $\Delta E$, energy difference between $x$ and $N(x)$ \label{alg:dadelta}
\ForEachParallel{variable $i$ in $x$}
\State propose a flip using $\Delta E_i - \Eoffset$
\sIf{acceptance criteria is satisfied}{record flip}
\EndFor
\If{at least one flip meets the acceptance criteria} 
    \State choose one flip at random from recorded flips
    \State set $x$ as solution corresponding to the selected flip
    \State $\Eoffset \assign 0$
\Else
\State $\Eoffset \assign \Eoffset + \offsetIncreaseRate$ \label{offset}
\EndIf
\EndFor
\State \Return best solution found 
\end{algorithmic}
\end{algorithm}

In Alg.~\ref{alg:da1}, we set the QUBO matrix $Q$ to be solved, the maximum number of iterations allowed ($\MaxIterations$), initial temperature ($\delta_0$), final temperature ($\delta_f$), temperature decay ($\xi$) and the offset increase rate parameter ($\offsetIncreaseRate$), which is used to escape local optima. We use a simple cooling scheme presented in line \ref{alg:dacool} to reduce the temperature from $\delta_0$ to $\delta_f$.

The DA architecture is designed to run on Fujitsu's specialised hardware \citep{MatTakMiyShi2020digital} (in a similar manner that some algorithms are designed to run on Graphical Processing Units),
such that all neighbouring solutions are explored in parallel and in constant time regardless of the number of neighbours. This approach significantly improves acceptance probabilities when compared to the classic SA algorithm \citep{aramon2019physics}. The DA algorithm does not completely evaluate a solution, rather, it calculates the energy difference between any solution $x$ and all its $m$ neighbours, $N(x)$. In the CPU implementation used in this study, we calculate the energy difference ($\Delta E$) between $x$ and all its neighbors $N(x)$ as

\begin{align}
\Delta E &=  \Biggl\{ 
P_{ii} \cdot (1-2x_i) + \sum_{\substack{j=1\\ j \neq i}}^{m} P_{ji} \cdot (1-2x_i) \cdot x_i, \;\;\; \forall i \in [1,m] \Biggr\} \nonumber\\
\label{eq:cale}
P &=  Q + Q^T, \ P_{ii} = P_{ii} - Q_{ii}\ \forall\ i \in [1,m] \;,
\end{align}
where $Q$ and $m$ are used to denote the QUBO matrix and its size, respectively. $P$ is the sum of $Q$ and the transpose $Q$ without the linear terms. $P$ only needs to be computed once, regardless of the number of iterations or runs.

To estimate $\Delta E$ (Eq.~\ref{eq:cale}), the decision to add or subtract the linear terms (i.e. $i=j$) is made depending on the variable $x_i$ being a $0$ or $1$. The quadratic terms (i.e. $i\neq j$) are deducted only if $x_i=1$. With this simple method, we can estimate $\Delta E$ for all neighbours of $x$ in one go. The energy (constrained objective function) of a solution $y$ that is the $i^\text{th}$ neighbour of $x$ is therefore $E(y) = E(x) + \Delta E_i$.

A new solution $y$ is accepted if its $\Delta E$ value is negative (i.e there is an improvement in the energy). It can also be accepted if $y$ is worse but within the acceptance probability. The acceptance probability is calculated as
%
\begin{align}
\label{eq:paccepts}
    \Pr(y \text{ is accepted}) = \exp(\min\{0, -(\Delta E_i -  \Eoffset)/\text{$\delta$}\})\;.
\end{align}

The probability of accepting a solution $y$, which is the $i^\text{th}$ neighbour of a current solution $x$, depends on the quality of the solution compared to $x$ ($\Delta E_i$), how long the search has been trapped in local optima ($\Eoffset$) and the current temperature $\delta$. Higher temperatures mean more exploration and higher probability of accepting degrading solutions (larger positive $\Delta E_i$ values). $\Eoffset$ is set to 0 when the algorithm is not trapped, but increases by $\beta$ for each iteration that the algorithm is not able to escape a local optima.

\subsection{Scalarisation-Based DA Algorithm}
\label{sec:sbda1}
The SB-DA algorithm is described in Alg. \ref{alg:sbda}. In addition to parameters ($\MaxIterations$, $\delta_0$, $\delta_f$, $\xi$, $\offsetIncreaseRate$) used by the DA, SB-DA requires additional inputs, which are $R$, $S$, $G$ and $\Gamma$. QUBO matrices $R$, $S$ and $G$ represent the first objective, second objective and constraint functions, respectively, while $\Gamma$ is a set of scalarisation weights.

In line \ref{line:costq} of Alg. \ref{alg:sbda}, $C$ is the QUBO matrix that represents the aggregated cost function. Penalty weight $\alpha$ (line \ref{line:alpha}) is set using method presented in Eq. \eqref{eq:wvermaa}. 

\begin{align}
\label{eq:vermap}
w = \max\Biggl\{& -C_{ii} - \sum_{\substack{j=1\\ j \neq i}}^{m}  \min\{C_{ij}, 0\}\ \forall\ i \in  \left [ 1,m \right ], C_{ii}
+ \sum_{\substack{j=1\\ j \neq i}}^{m} \max\{C_{ij}, 0\}\  \forall\ i \in  \left [ 1,m \right ] \Biggr\} \\
\label{eq:wvermaa}
\alpha = \frac{w}{2}&
\end{align}

Equation \eqref{eq:vermap} was originally proposed in \citep{VerLew2020penalty}, where $w$ is an upper bound of the difference in $C$ that can be achieved by either turning a bit off (from 1 to 0) or on (from 0 to 1). Verma and Lewis \citep{VerLew2020penalty} suggested that $w$ can be further reduced using information from the constraint function. Equation \eqref{eq:wvermaa} was proposed in \citep{Ayodele2022penalty} for permutation problems represented as two-way one-hot, given that the minimum constraint function value of any infeasible solution (i.e. solutions that are one flip away from feasibility) is 2.

The SB-DA algorithm makes multiple calls to the DA (Alg. \ref{alg:da1}) using different $Q$ matrices ($Q$ is a weighted aggregate of the QUBOs representing the objectives and constraints) and returns a set of non-dominated solutions.

\begin{algorithm}[t]
\caption{SB-DA Algorithm}\label{alg:sbda}
\begin{algorithmic}[1]
\Require $R$, $S$, $G$, $\MaxIterations$, $\delta_0$, $\delta_f$, $\xi$, $\offsetIncreaseRate$, $\Gamma$
\State $A \assign \emptyset$\Comment{Initialise archive}
\ForEach{$\gamma \in \Gamma$}
\State $C \assign\ \gamma \times R + (1- \gamma) \times S$  \label{line:costq}
\State estimate $\alpha$ from $C$ \label{line:alpha}
\State $Q \assign C + \alpha \times G$ 
\State $x \assign$ execute the DA with parameters $Q$, $\MaxIterations$, $\delta_0$, $\delta_f$, $\xi$, $\offsetIncreaseRate$
\State add $x$ to $A$
\EndFor
\State \Return all non-dominated solutions from archive $A$ 
\end{algorithmic}
\end{algorithm}

\subsection{Multi-objective Digital Annealer Algorithm}
In this study, we adapt the DA algorithm as a multi-objective algorithm where the energies corresponding to all the objective functions $(E_1(x), E_2(x))$ are simultaneously optimised. The relationship between any two solutions $x$ and $y$ can be defined in any of the following ways. 
\begin{enumerate}
\item $x$ is equivalent to $y$ (denoted by $x \sim y$), if $E_i(x) = E_i(y)\ \forall\ i \in  \{ 1,2  \}$
\item  $x$ dominates $y$ (denoted by $x \prec y$), if $E_i(x)\leq E_i(y)\ \forall\ i \in \{ 1,2  \}$ and $\lnot(x \sim y)$ 
 \item $x$ and $y$ are mutually non-dominated (denoted by $x \nondom y$), if $\exists\ i, j \in  \{ 1,2  \}$ such that $E_i(x) < E_i(y)$ and $E_j(x) > E_j(y)$
  \end{enumerate}

\begin{algorithm}[htb]
\caption{MDA Algorithm}\label{alg:moda0}
\begin{algorithmic}[1]
\Require $Y$, $Z$, $\MaxIterations$, $\delta_0$, $\delta_f$, $\xi$, $s$
\State Initialise archive $A$ with maximum size $s$
\State $x \assign$ a random binary solution of length $m$
\State $\delta  \assign  \delta_0$
\For {iteration = 1 to $\MaxIterations$ }
\sIf{$\delta > \delta_f$}{$\delta \assign  \delta \cdot( 1-\xi)$}
\State calculate $\Delta E$, energy difference between $x$ and $N(x)$ for all objectives
\ForEachParallel{variable $i$ in $x$}
\State propose a flip using $\Delta E_{ik}\ \forall\ k \in [1,2]$  \label{MDAe}
\sIf{acceptance criteria is satisfied}{record flip}
\EndFor
\If{at least one flip meets the acceptance criteria} 
    \State choose one flip at random from recorded flips \label{MDAflip}
    \State set $x$ as solution corresponding to the selected flip
    \State update($A$, $x$)
\Else
\State $x \assign$ a randomly selected solution from $A$ \label{mdaescape}
\EndIf
\EndFor
\State \Return all non-dominated solutions from archive $A$ 
\end{algorithmic}
\end{algorithm}

The MDA algorithm proposed in this work is presented in Alg.~\ref{alg:moda0}.
Similar to the DA, MDA uses temperature parameters $\delta_0$, $\delta_f$ and $\xi$ as well as the stopping criteria ($\MaxIterations$). Additional inputs used are QUBO matrices $Y$ and $Z$ as well as archive size $s$. QUBO matrices $Y$ and $Z$ are derived from $R$, $S$ and $G$ that are QUBO matrices representing the first objective, second objective and constraint functions respectively (Eq. \ref{eq:wqubom}):
\begin{align}
\label{eq:wqubom}
Y = R + \alpha_1 \times G, \quad  Z = S + \alpha_2 \times G
\end{align}

 Penalty weights $\alpha_1$ or $\alpha_2$ are derived by substituting $R$ or $S$ for $C$ in Eq. \eqref{eq:vermap}.
 
Like the DA, neighbours $N(x)$ of a solution $x$ are evaluated by exploring the difference in energy $\Delta E$. In this case however, the difference is computed using QUBOs ($Y$ and $Z$, Eq. \ref{eq:wqubom}) representing each objective. $\Delta E_{ik}$ (line \ref{MDAe}) denotes the energy difference between $x$ and the $i^\text{th}$ neighbour of $x$ w.r.t. the $k^\text{th}$ objective. $\Delta E_{i1}$ (or $\Delta E_{i2}$) is estimated by substituting $Y$ (or $Z$) for $Q$ in Eq. \eqref{eq:cale}.

The difference between the SB-DA algorithm used in this study and the proposed MDA are: (1) SB-DA solves a sequence of QUBOs independently while the MDA solves two QUBOs simultaneously, they therefore require different acceptance criteria, (2) multiple pre-processing of QUBOs are required in the SB-DA,once for each scalarisation weight in $\Gamma$, while pre-processing is done only once for MDA and (3) the methods of escaping local optima are different.

\paragraph{Acceptance Criteria} Two acceptance criteria (strict and lenient), adapted from \citep{amine2019multiobjective},  are used in this study. Given that $y$ is the $i^\text{th}$ neighbour of $x$, the strict acceptance criteria (Eq. \ref{eq:pacceptms}) uses the product of the probabilities of accepting a solution $y$ within the context of each individual objective. The lenient criteria (Eq. \ref{eq:pacceptml}) uses the highest probability of accepting a solution $y$ with respect to any of the objectives.
\begin{align}
    \label{eq:pacceptms}
    \Pr(y \text{ is accepted}) & = \prod\nolimits_{k =1}^{2}{\Pr}^k(y \text{ is accepted}) \\
    \label{eq:pacceptml}
    \Pr(y \text{ is accepted}) & = \max\nolimits_{k =1}^{2}{\Pr}^k(y \text{ is accepted})  \\
    \text{where }  {\Pr}^k&(y \text{ is accepted})  = \exp(\min\{0, -\Delta E_{ik}/\text{$\delta$}\})\;.\notag
\end{align}

The lenient acceptance rule provides the certainty that non-comparable or dominating solutions are accepted while the strict acceptance rule allows for deeper exploration \citep{amine2019multiobjective}.

\paragraph{Escaping Local Optimal} The parameter $\offsetIncreaseRate$, used to escape local optimal in the DA, is not used in the MDA because the latter is able to explore other areas of the search space by switching to another solution in the archive (see line \ref{mdaescape}).

\paragraph{Updating the Archive} In the MDA, an archive is used to store multiple non-dominated solutions. A fixed-size archive of size $s$ is used in this study. At each iteration, one of the neighbouring solutions that meet the acceptance criteria is randomly selected as a potential solution to be added to the archive (line \ref{MDAflip}). We explore two methods of updating the archive: A more explorative method and a more exploitative method.

\begin{algorithm}[htb]
\caption{update($A$, $x$): Explorative Method}\label{alg:explore}
\begin{algorithmic}[1]
\Require $A$: archive, $x$: new solution, $s$: maximum size
    \State dominates $\assign$ \false
    \ForEach{$y \in A$}
    \If{$x \prec y$}
      \State dominates $\assign$ \true
      \State \textbf{break}
    \EndIf
    \EndFor
    \If{dominates}
    \State replace $y$ with $x$ in $A$
    \ElsIf{$|A| < s$} \State add $x$ to $A$
    \EndIf
\end{algorithmic}
\end{algorithm}

In the explorative method (Alg. \ref{alg:explore}), if an accepted solution $x$ dominates any solution in the archive, then $x$ replaces the first of the dominated solutions. However, if the accepted solution $x$ does not dominate any solution but the archive is not full, then such solution $x$ is also added to the archive.

\begin{algorithm}[htb]
\caption{update($A$, $x$): Exploitative Method}\label{alg:exploit}
\begin{algorithmic}[1]
\Require $A$: archive, $x$: new solution, $s$: maximum size
\State $B \assign \{ y \in A \mid x \prec y \}$ 
\If{$\nexists\ z \in A,  z \prec x$}
    \If{$|B| > 0$}
    \State select a solution $y$ from $B$ at random
    \State replace $y$ with $x$ in $A$
    \ElsIf{$|A| < s$} \State add $x$ to $A$
    \EndIf
\EndIf
\end{algorithmic}
\end{algorithm}

While the explorative method updates the archive with solutions that meet the acceptance criteria, such solutions must be non-dominated for it to be added to the archive in the more exploitative method (Alg. \ref{alg:exploit}). The non-dominated solution $x$ is added to the archive either by replacing a dominated solution selected at random or by filling a new position in the archive (if the size of the archive is less than the maximum size $s$).

\section{Experimental Settings}
\label{sec:exps}
In this section, the instances used in this study are described. The parameter settings and performance measures are also defined.

\subsection{Bi-Objective QAP Instances}
\label{sec:bqi}
We use publicly available bi-objective QAP instances~\citep{LopPaqStu05:jmma,PaqStuLop07metaheuristics}.\footnote{Available from \url{https://eden.dei.uc.pt/~paquete/qap/}, equivalent QUBOs available from \url{https://github.com/mayoayodelefujitsu/QUBO_Biobjective-Quadratic-Assignment-Problem}} The benchmark consists of instances classed as structured or unstructured, with 25, 50 or 75 locations (and facilities) and different levels of correlations between their objectives. It also consists of best known objectives (relating to the Pareto sets) for the problem sets. Details of how the problems are generated can be found in \citep{KnoCor2003emo}.

In this study, we use structured instances with varying numbers of locations/facilities $n \in\{25, 50\}$ and different levels of correlations $\rho \in\{-0.75, 0, 0.75\}$ between the objectives. Instances are named as qapStr.$n$.$X$.$i$, where $n$ is the instance size, $X \in \{\text{n75, 0, p75}\}$ denotes the correlation, and $i=\{1,2,3\}$ is a numerical identifier to distinguish instances of the same size and correlation. Instances that have negative correlation between their objectives (\mbox{$\rho = -0.75$}) simulate problems that have conflicting objectives. Instances with no correlation between the objectives ($\rho = 0$) simulate scenarios where the objectives are independent of each other. Where there is positive correlation (\mbox{$\rho = 0.75$}), we expect that minimising one objective should often lead to the minimisation of the other objective. 

\subsection{Parameter Settings}
The parameter settings used in this study for SB-DA and MDA are presented in Table \ref{tb:param}. These values were chosen based on preliminary experiments. Each algorithm configuration was run on all problem instances, each run was independently repeated 20 times with unique random seeds.

Furthermore, to prevent bias towards one of the objectives, we consider normalisation. We set the largest QUBO coefficient representing any objective QUBO $R$ and $S$ to $2^{23}$. These are inputs in SB-DA (Alg. \ref{alg:sbda}) and used to compute $Y$ and $Z$ (Eq. \ref{eq:wqubom}), which are inputs in the MDA (Alg. \ref{alg:moda0}). The maximum value ($2^{23}$) was derived empirically as we found that it was large enough to capture the information in the original QUBO. This was true for all the instances used in this study. We compare the quality of solutions returned when normalised or non-normalised QUBOs are used (Section \ref{sec:normal}).

We compare results for SB-DA and MDA using non-dominated solutions from all 20 runs.

\begin{table}[t]
\caption{SB-DA and MDA parameters and their settings.\label{tb:param}}
\centering
\resizebox{0.5\columnwidth}{!}{  
\begin{tabular}{rcc}
\toprule
\textbf{Parameter} & \textbf{SB-DA values}  & \textbf{MDA values} \\\midrule
Initial temperature $\delta_0$ & $10^9$ & $10^9$ \\
Final temperature $\delta_f$ & $10^4$ & $10^4$\\
Number of iterations  $\MaxIterations$ & $0.25 \cdot m^2$ & $0.25 \cdot m^2$ \\
Number of runs & $20$ & $20$ \\
Offset increase rate $\offsetIncreaseRate$ & $\delta_0/ (0.25 \cdot m^2)$ & N/A\\ 
Temperature decay $\xi$ & 0.001 & 0.001\\
Scalarisation weights $\Gamma$ & $\{0.0, 0.1, 0.2,\dots 1.0\}$ & N/A\\
Size of archive $s$ & N/A & $m$ \\ \bottomrule
\end{tabular}}
\end{table}

\subsection{Performance Criteria}
Several performance measures have been proposed for comparing multi-objective optimisers. Following best practices~\citep{ZitKnoThi2008quality}, we use the Empirical Attainment Function (EAF) and the hypervolume metric in this study.

The EAF of an algorithm gives the probability, estimated from multiple runs, that the non-dominated set produced by a single run of the algorithm dominates a particular point in the objective space.  The visualisation of the EAF~\citep{Grunert01} has been shown as a suitable graphical interpretation of the quality of the outcomes returned by local search methods. The visualisation of the differences between the EAFs of two alternative algorithms indicates how much better one method is compared to another in a particular region of the objective space~\citep{LopPaqStu09emaa}. The EAF visualisations were done using the \texttt{eaf} \textsf{R} package.\footnote{\url{http://lopez-ibanez.eu/eaftools}}

The hypervolume quality metric measures the area in the objective space that is dominated by at least one of the points of a non-dominated set and bounded by a given reference point that must be dominated by all points under comparison. Larger hypervolume values indicate that the non-dominated solutions are closer to the Pareto front. In this study, we calculate the hypervolume using the \texttt{pymoo} Python library \citep{pymoo}.
Before computing the hypervolume, we normalise the energies of all the solutions generated in the study (together with the known energies for the Pareto optimal solutions) to the range $[1, 2]$. The reference point is set as $(2.1,2.1)$, hence, the maximum hypervolume possible is $1.21$.

\section{Results}
\label{sec:results}
 In this section, we first assess the effect of normalisation in the algorithm. We also compare the two acceptance criteria (lenient vs strict) and archiving methods (explore vs exploit) in MDA. Finally, we compare the quality of results produced by the SB-DA with those produced by MDA.

\subsection{Effect of QUBO Normalisation}
\label{sec:normal}

\begin{figure}[t]
  \centering%
    \includegraphics[width=0.85\columnwidth]{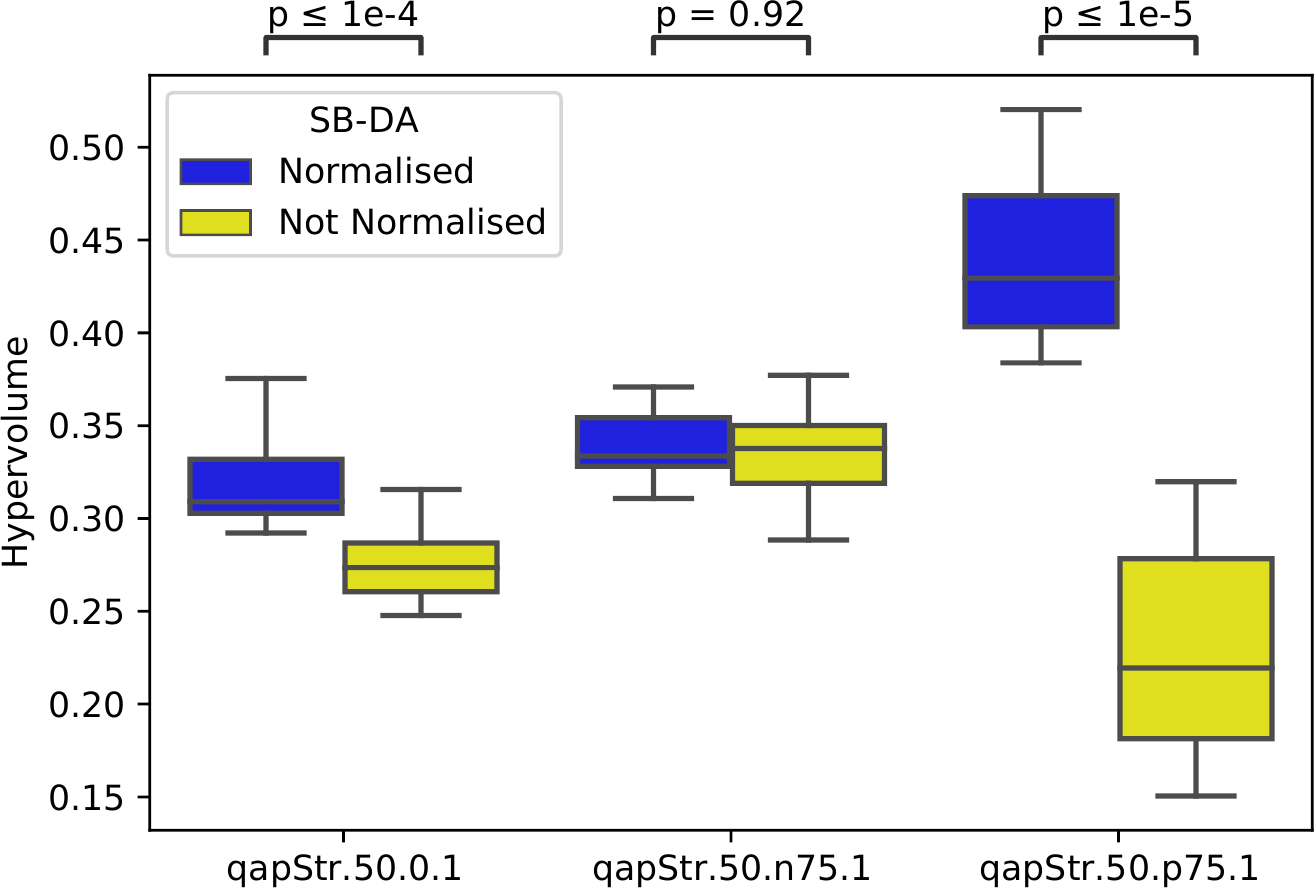}
  \caption{SB-DA: effect of QUBO normalisation}
   \label{fig:das}
\end{figure}

\begin{figure}[t]
  \centering%
  \includegraphics[width=0.85\columnwidth]{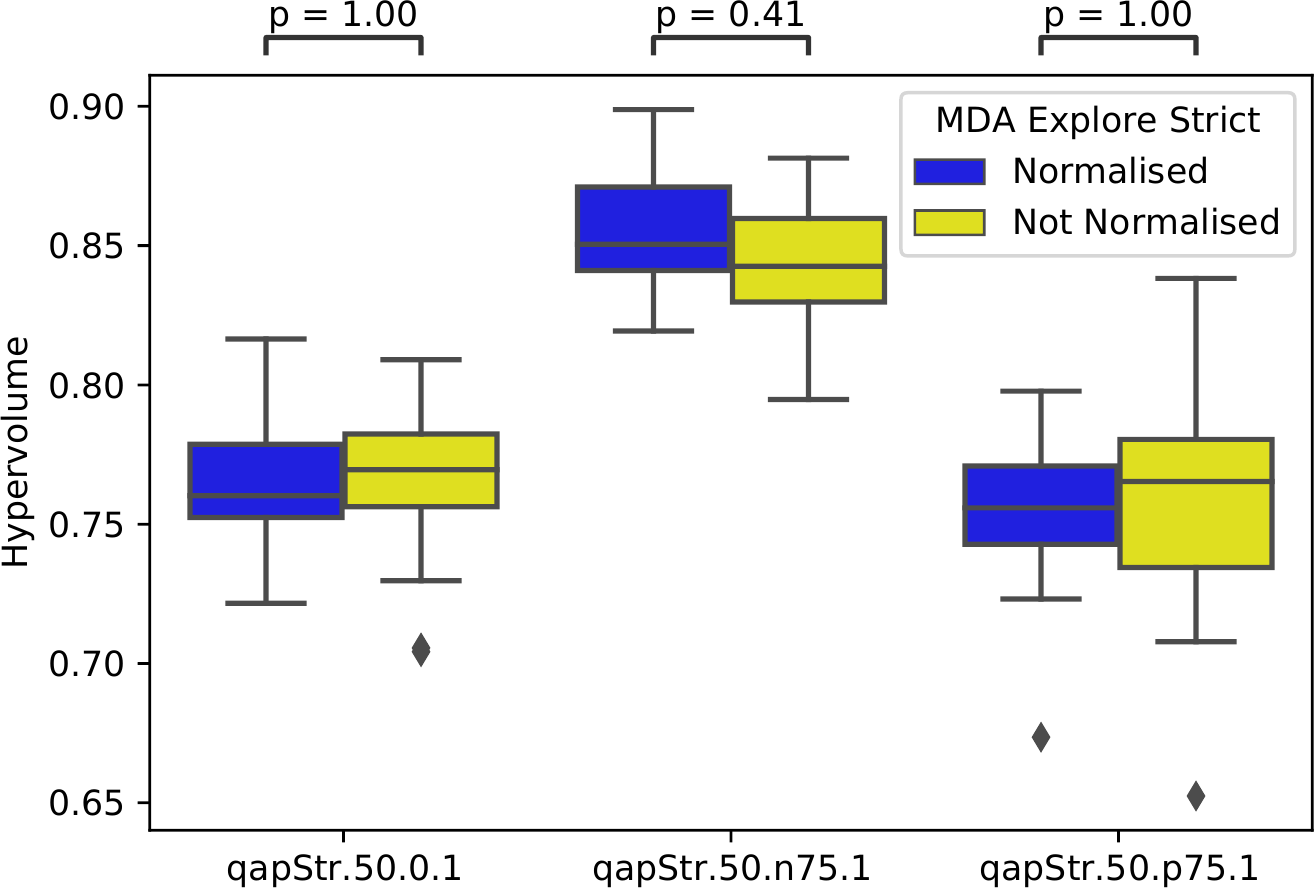}
  \caption{MDA: effect of QUBO normalisation}
   \label{fig:mdasb}
\end{figure}

Figures \ref{fig:das} and \ref{fig:mdasb} present the hypervolume of the non-dominated solutions across 20 runs of the SB-DA and MDA respectively. We use the paired $t$-test to test for statistical significance (the corresponding p-values are shown in the plot). While Figures \ref{fig:das} and \ref{fig:mdasb} only show results for the first instance of each set in qapStr.$50$ (details of problems set are presented in Section \ref{sec:bqi}), results for all nine instances consisting of 25 locations and facilities (qapStr.25) as well as all nine instances consisting of 50 locations and facilities (qapStr.$50$) are presented in the supplementary material. 

In SB-DA, the normalised QUBOs led to significantly better (\mbox{p-value} $\leq$ 0.05) hypervolume values on all nine qapStr.25. We also show similar results on qapStr.50 instances with no correlation (qapStr.$50$.$0$) or positive correlation (qapStr.$50$.$p75$) between the objectives. However, for instances with negative correlation between the objectives (qapStr.$50$.$n75$), the SB-DA with normalised QUBOs did not present significantly better results on two of three instances when compared to using QUBOs that are not normalised (first instances shown in Fig. \ref{fig:mdasb} and the other instances shown in the supplementary material). 

We use the MDA with more explorative archiving method and strict acceptance criteria in this section because these are the best settings for the MDA (Section \ref{mdaadc}). In the MDA, there are no statistically significant (p-value > 0.05) differences between using the MDA with QUBOs that are either normalised or not (Figure \ref{fig:mdasb}).

For consistency, results presented in the following sections relating to SB-DA and MDA use normalised QUBOs.

\subsection{MDA: Assessing Algorithm Design Choices}
\label{mdaadc}
\begin{figure}[t]
  \centering%
  \foreach \z in {25scaled,50scaled} {
    \includegraphics[width=0.6\columnwidth]{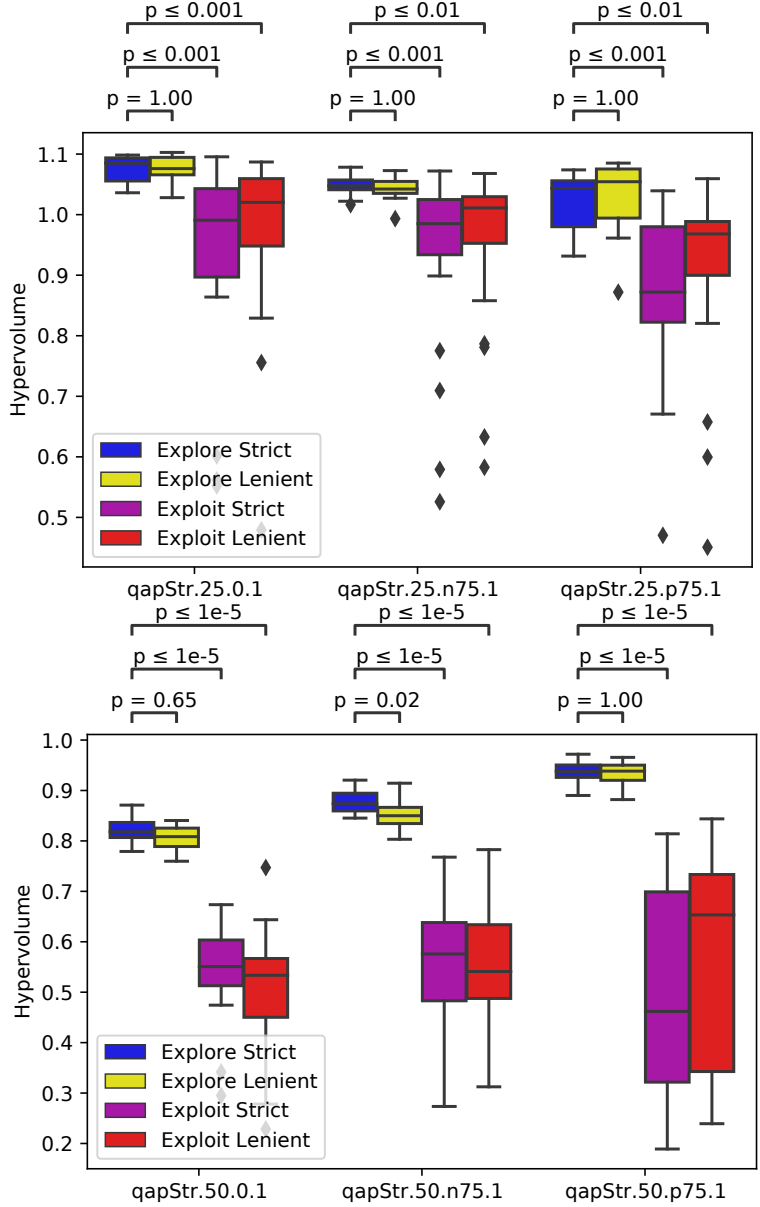}\newline\vspace{1ex}}
  \vspace*{-1em}
  \caption{Comparing archive update methods and acceptance criteria in the MDA}
   \label{fig:mdas}
\end{figure}

We compare the more explorative archive update method with the more exploitative archive update method using different acceptance criteria (strict and lenient). We test for significance in the difference between mean hypervolume values (20 independent runs) using the Mann-Whitney-Wilcoxon two-sided test with Bonferroni correction. We show the corresponding p-values in Figure \ref{fig:mdas}.

Figure \ref{fig:mdas} shows that the MDA using a more explorative method of updating the archive presented better results (larger hypervolume) than the more exploitative method. The difference between the performance (hypervolume) of the MDA with strict and lenient criteria within the context of the more explorative method (\textit{Explore Lenient} vs \textit{Explore Strict}) are not of statistical significance (p-value > 0.05). 

Using the exploitative method of updating the archive (\textit{Exploit Strict} and \textit{Exploit Lenient}) leads to larger variance between the hypervolume across all 20 runs compared to the explorative method. 

\begin{figure}[h]
\begin{center}
   \includegraphics[width=0.6\columnwidth]{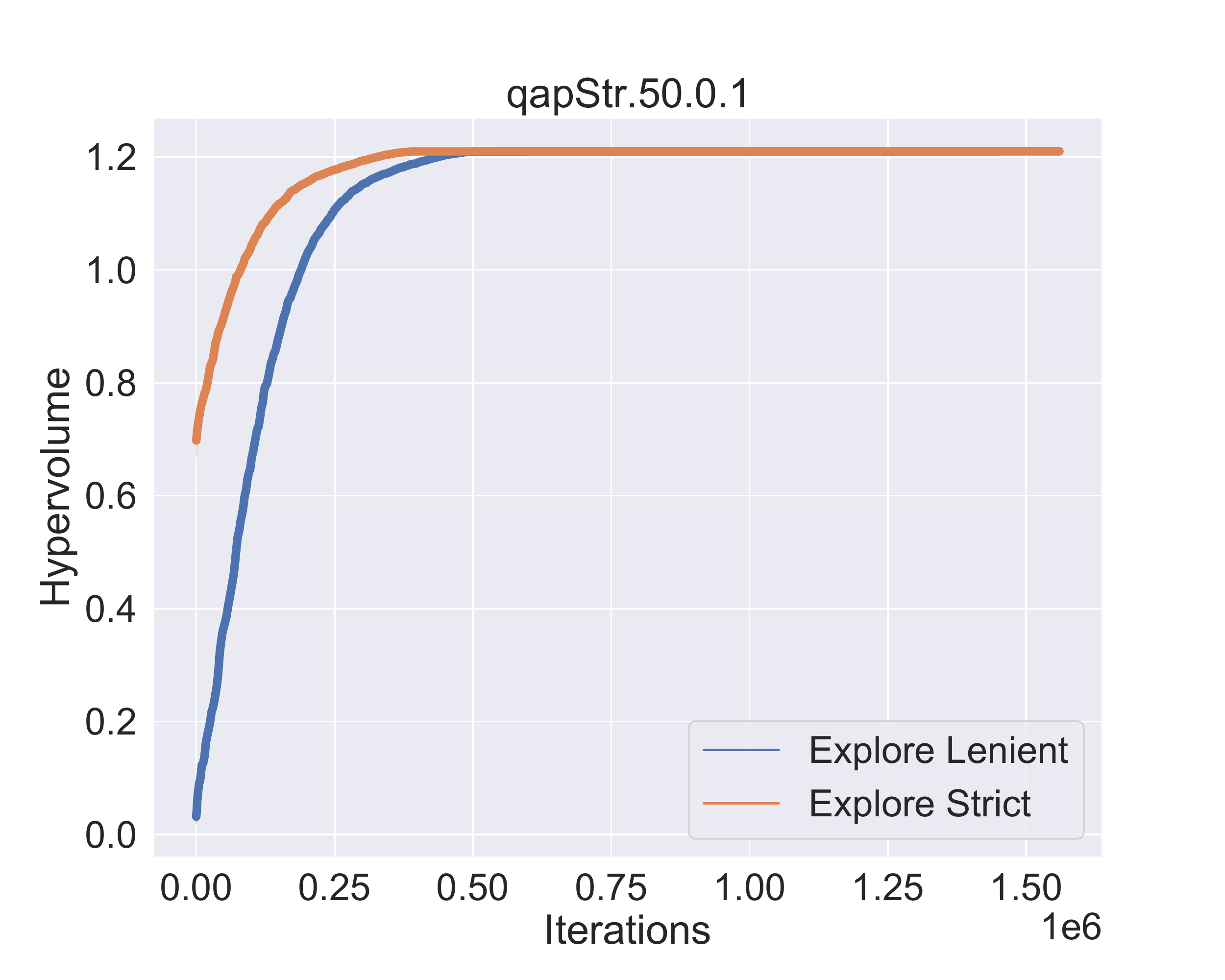}
  \caption{Comparing archive acceptance criteria in the MDA}
   \label{fig:mdasl}
   \end{center}
\end{figure}

In Figure \ref{fig:mdasl}, we examine the acceptance criteria (with the more explorative archiving method) in more detail looking at the hypervolume recorded at each iteration. We show that the strict criteria converges faster than the lenient criteria on qapStr.50.0.1. Similar results are observed on all instances (qapStr.25 and qapStr.50), results are presented in the supplementary material.

\subsection{Comparing SB-DA and MDA}
To understand the possible advantage of extending the DA as a multi-objective algorithm, we compare the SB-DA with the proposed MDA (using more exploratory archive method and the strict acceptance criteria).

\begin{figure}[h]
  \centering%
    \includegraphics[width=0.7\columnwidth]{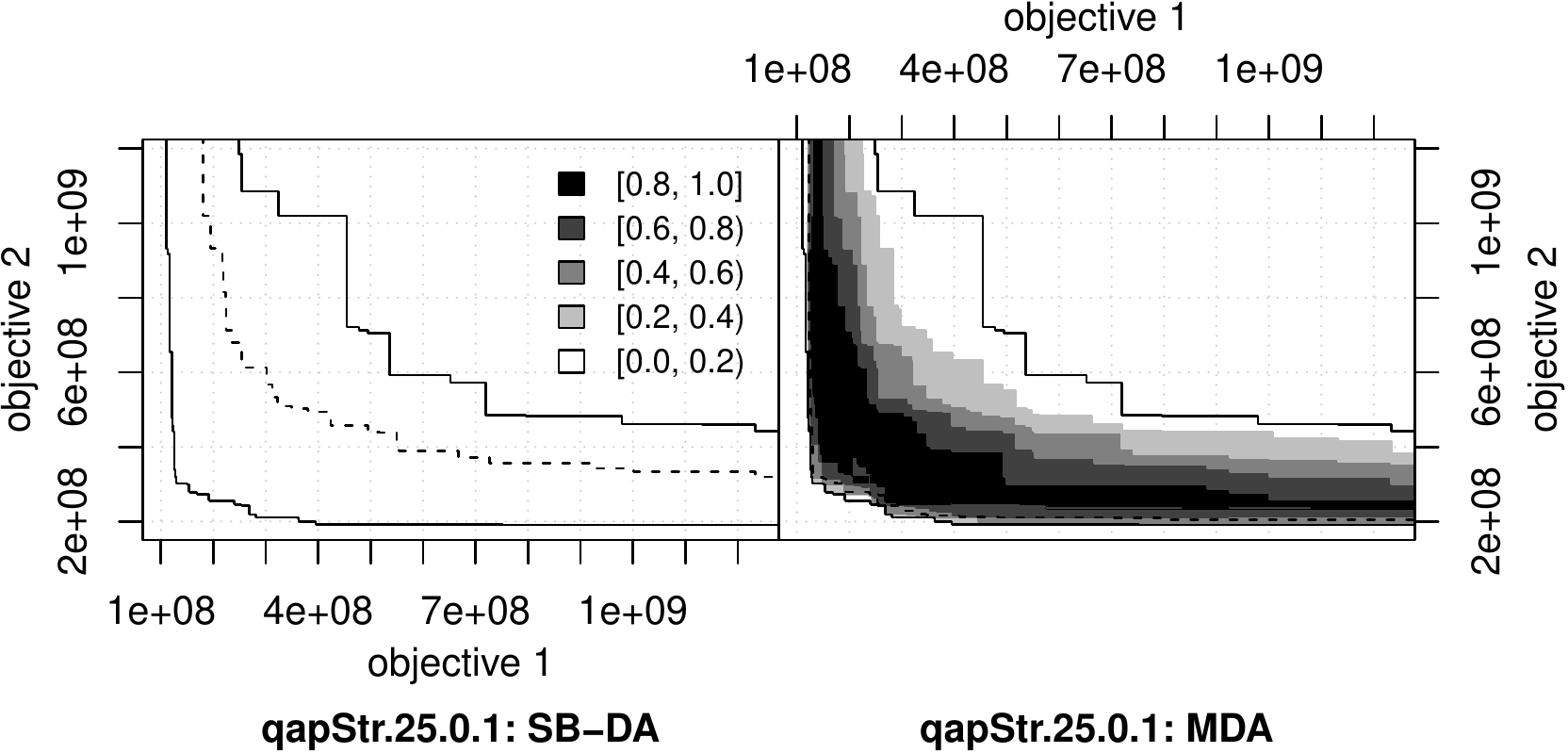}
   \includegraphics[width=0.7\columnwidth]{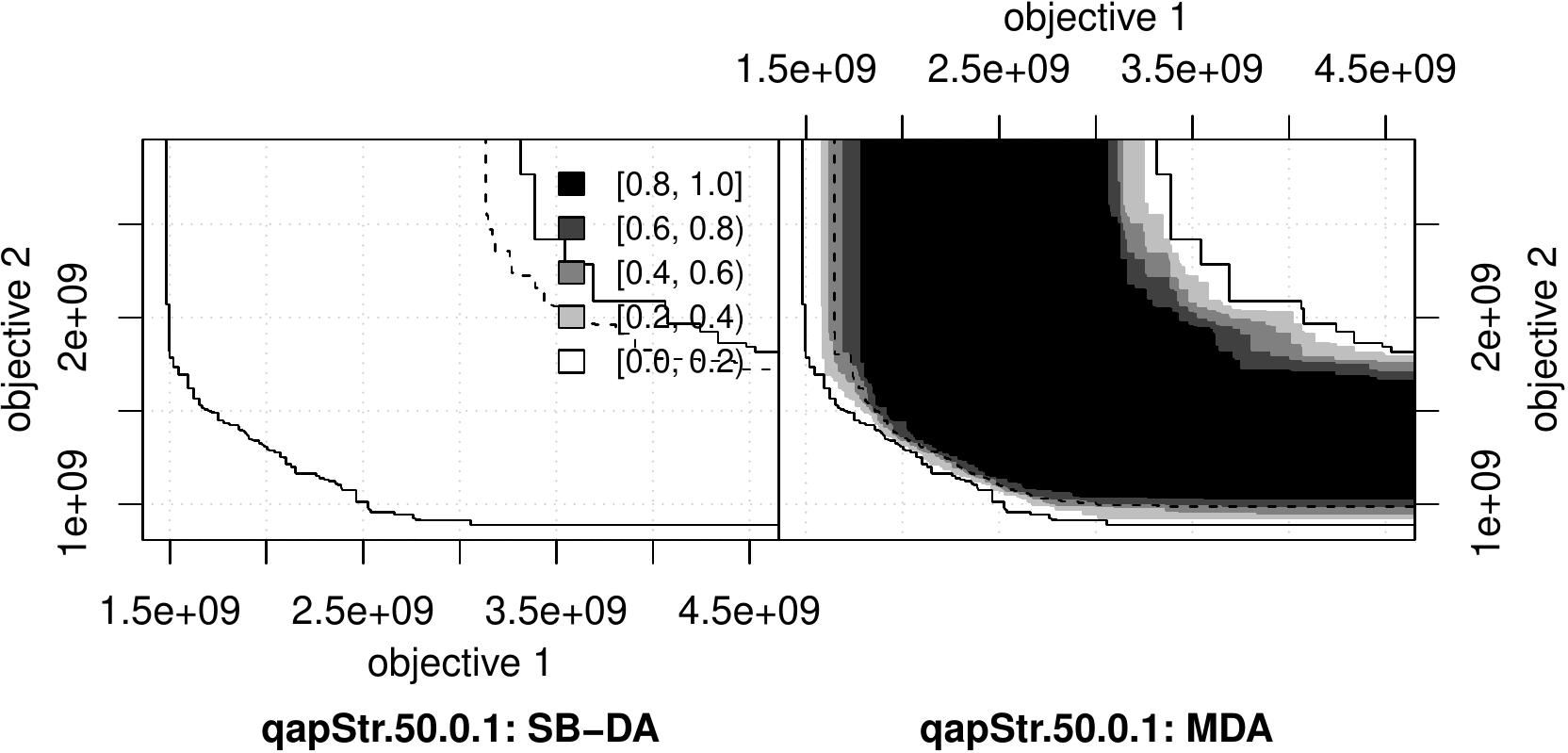}
  \caption{Comparing SB-DA with MDA on bi-objective QAP instances}
   \label{fig:damda}
\end{figure}

Figure \ref{fig:damda} shows the location of the EAF differences between the outcomes of the SB-DA and MDA with respect to their corresponding EAFs. The darker regions are the locations where the corresponding algorithm is better than the other algorithm. We see that the MDA is able to reach better non-dominated solutions (shown by the darker regions) across the front, than the SB-DA. Larger darker regions are shown in the nine qapStr.50 instances compared to the qapStr.25 instances (results are shown in the supplementary material). There is no region in the objective space where the SB-DA outperforms the MDA on all eighteen instances. 

We also compare the number of non-dominated solutions returned by the SB-DA and MDA in Table \ref{tb:arcs}. The SB-DA returned 3-5 non-dominated solutions while the MDA returned 24-98 non-dominated solutions. This is not unexpected as the maximum number non-dominated solutions that SB-DA can return in one run is the size of the scalarisation weights ($\Gamma$) while MDA can return up to $s=m$ solutions.

\begin{table}[t]
\caption[Number of non-dominated solutions in the archive]{Number of non-dominated solutions in the archive (mean \textpm{} std. dev.).\label{tb:arcs}}
\begin{center}
\begin{tabular}{*{7}{r}}
\toprule
 & \multicolumn{3}{c}{qapStr.25.} & \multicolumn{3}{c}{qapStr.50.} \\  \cmidrule{2-4} \cmidrule(l){5-7} 
 & 0.1 & n75.1 & p75.1 & 0.1 & n75.1 & p75.1 \\ \cmidrule{1-4}\cmidrule(l){5-7} 
SB-DA & 4 \textpm 1 &  5 \textpm 1 &  4 \textpm 1 & 5 \textpm 1 &  5 \textpm 1  &  3 \textpm 1 \\ 
MDA &  36 \textpm 9 & 44 \textpm 8 &  24 \textpm 7 & 81 \textpm 20 & 98 \textpm 25 &  70 \textpm 16 \\ \bottomrule
\end{tabular}
\end{center}
\end{table}

\begin{table}[t]
\caption{Computation Times in Seconds (mean \textpm{} std. dev.).\label{tb:tts}}
\begin{center}
\begin{tabular}{@{}*{7}{r}@{}}
\toprule
 & \multicolumn{3}{c}{qapStr.25.} & \multicolumn{3}{c}{qapStr.50.} \\ \cmidrule{2-4} \cmidrule(l){5-7} 
 & 0.1 & n75.1 & p75.1 & 0.1 & n75.1 & p75.1 \\ \cmidrule{1-4} \cmidrule(l){5-7} 
SB-DA& 47 \textpm 0 & 51 \textpm 1 & 47 \textpm 0 & 1,967 \textpm 17 & 1,963 \textpm 20  & 1,976 \textpm 18 \\ 
MDA & 8 \textpm 1 & 8 \textpm 1 & 8 \textpm 1 & 427 \textpm 14 & 445 \textpm 13 & 458 \textpm 17 \\ \bottomrule
\end{tabular}
\end{center}
\end{table}

We further compared the speed of arriving at the non-dominated solutions returned by the SB-DA and MDA in Table \ref{tb:tts}. All executions were performed on a machine equipped with Ubuntu 18.04, Intel Xeon Gold 5218 CPU @ 2.30GHz, and 192GB Memory. Results show that the MDA is an average of 4-6 times quicker than SB-DA. We conclude that MDA can present better quality and larger quantity of non-dominated solutions faster than SB-DA. We note that the DA that uses specialised hardware is orders of magnitude faster than the CPU implementation used in this study.

\section{Conclusion}
\label{sec:con}
In this study, we presented methods of solving the bi-objective QAP formulated as QUBO. We explored scalarisation as well as Pareto optimisation. In the scalarisation approach, we assessed the effect of QUBO normalisation such that QUBO representing different objectives are put on the same scale before they are aggregated. We use the existing DA algorithm, which is similar to SA, to solve aggregated QUBOs as a single-objective problem. Furthermore, we also explored methods of extending the existing single-objective DA to a multi-objective algorithm. We particularly focused on methods of updating the archive as well as solution acceptance criteria in the MDA. We show that more promising non-dominated solutions can be attained using the MDA compared to running the single-objective DA with multiple scalarisation weights. We also show that these more promising solutions can be reached by the MDA in significantly shorter time. Moreover, uploading multiple large QUBOs may become expensive when the DA implemented on specialised hardware is used due to the time needed to upload such QUBOs to the DA.

This work has multiple future directions. First, we plan to assess the sensitivity of parameters of the algorithms. Second, we plan to investigate better ways of aggregating QUBOs for multiple objectives. Third, we want to assess the performance of the proposed MDA on more problems. Finally, we wish to extend the proposed MDA for solving more than two objectives as well as explore other ways of handling exploration and archiving~\citep{DubLopStu2015ejor,LopKnoLau2011emo}.



\bibliographystyle{unsrtnat}
\bibliography{bib/abbrev,bib/journals,bib/authors,bib/biblio,bib/crossref,References}


\end{document}


\begin{abstract}
   This document supplements paper titled ``Multi-objective QUBO Solver for Bi-objective Quadratic Assignment Problem''. More results relating to all structured instances of the bi-objective QAP that consists of 25 and 50 locations (and facilities) are presented. Links to contents are provided for ease of navigation.
\end{abstract}

\newpage

\maketitle

\listoffigures
\listoftables
\newpage

\begin{figure}
\foreach \z in {0,n75,p75} {
   \includegraphics[width=0.7\columnwidth]{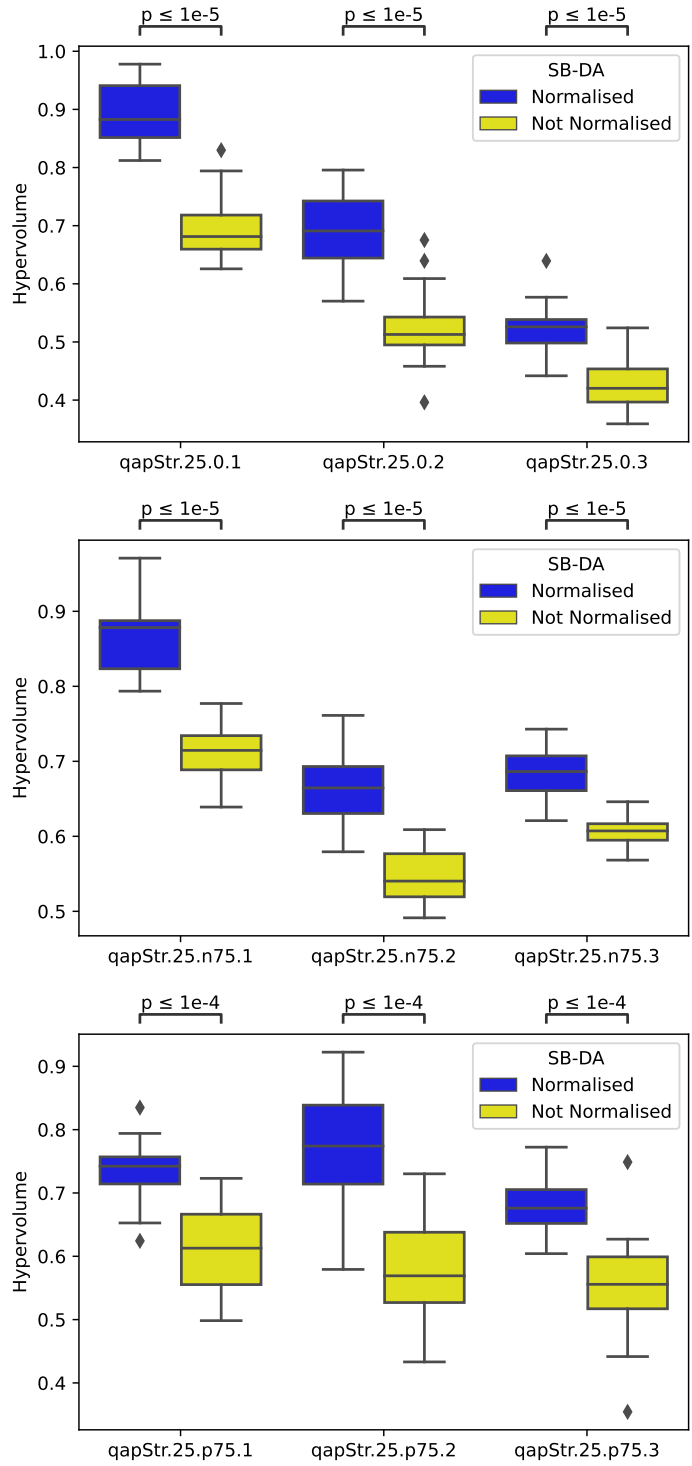}}
    \caption{SB-DA: assessing the effect of QUBO normalisation (qapStr.25)}
   \label{fig:dasa1}
\end{figure}

\begin{figure}
\foreach \z in {0,n75,p75} {
   \includegraphics[width=0.7\columnwidth]{graphics/boxplots/sqap50.\z.pdf}}
   \caption{SB-DA: assessing the effect of QUBO normalisation (qapStr.50)}
   \label{fig:dasa2}
\end{figure}

\begin{figure}
\foreach \z in {0,n75,p75} {
   \includegraphics[width=0.7\columnwidth]{graphics/boxplots/MSqap25.\z.pdf}}
  \caption{MDA: assessing the effect of QUBO normalisation (qapStr.25)}
   \label{fig:dasa3}
\end{figure}

\begin{figure}
\foreach \z in {0,n75,p75} {
   \includegraphics[width=0.7\columnwidth]{graphics/boxplots/MSqap50.\z.pdf}}
  \caption{MDA: assessing the effect of QUBO normalisation (qapStr.50)} 
   \label{fig:dasa4}
\end{figure}

\begin{figure}
\foreach \z in {0,n75,p75} {
   \includegraphics[width=0.7\columnwidth]{graphics/boxplots/mqap25.\z.pdf}}
  \caption{Comparing archive update methods and acceptance criteria in the MDA (qapStr.25)}
   \label{fig:mdas25aa}
\end{figure}

\begin{figure}
\foreach \z in {0,n75,p75} {
   \includegraphics[width=0.7\columnwidth]{graphics/boxplots/mqap50.\z.pdf}}
  \caption{Comparing archive update methods and acceptance criteria in the MDA (qapStr.50)} 
   \label{fig:mdas50aa}
\end{figure}

\foreach \z in {0,n75,p75} {
\begin{figure}
\foreach \y in {1Scaled,2Scaled,3Scaled} {
   \includegraphics[width=0.7\columnwidth]{graphics/generational_plots/ExploreqapStr.25.\z.\y.pdf} }
    \caption{Comparing archive acceptance criteria in the MDA (qapStr.25.\z)}
\end{figure}}

\foreach \z in {0,n75,p75} {
\begin{figure}
\foreach \y in {1Scaled,2Scaled,3Scaled} {
   \includegraphics[width=0.7\columnwidth]{graphics/generational_plots/ExploreqapStr.50.\z.\y.pdf} }
    \caption{Comparing archive acceptance criteria in the MDA (qapStr.50.\z)}
\end{figure}}

\foreach \y in {0,n75,p75} {
\begin{figure}
\foreach \z in {1,2,3} {
   \includegraphics[width=1\columnwidth]{graphics/eaf/qapStr.25.\y.\z_SONormalised_productExploreScaled.pdf}}
  \caption{Comparing SB-DA with MDA on bi-objective QAP instances (qapStr.25.\y)}
 \end{figure}
}

\foreach \y in {0,n75,p75} {
\begin{figure}
\foreach \z in {1,2,3} {
   \includegraphics[width=1\columnwidth]{graphics/eaf/qapStr.50.\y.\z_SONormalised_productExploreScaled.pdf}}
  \caption{Comparing SB-DA with MDA on bi-objective QAP instances (qapStr.50.\y)}
 \end{figure}
}

\foreach \y in {0,n75,p75} {
\begin{figure}
\foreach \z in {1,2,3} {
   \includegraphics[width=1\columnwidth]{graphics/eaf/qapStr.25.\y.\z_productExploitScaled_productExploreScaled.pdf}}
  \caption{Comparing Explore and Exploit Archiving Methods in MDA for bi-objective QAP instances (qapStr.25.\y)}
 \end{figure}
}

\foreach \y in {0,n75,p75} {
\begin{figure}
\foreach \z in {1,2,3} {
   \includegraphics[width=1\columnwidth]{graphics/eaf/qapStr.50.\y.\z_productExploitScaled_productExploreScaled.pdf}}
  \caption{Comparing Explore and Exploit Archiving Methods in MDA for bi-objective QAP instances (qapStr.50.\y)}
 \end{figure}
}

\begin{table}[b]
\caption{Number of Non-Dominated Solutions\label{tb:tts}}
\centering
\resizebox{0.7\columnwidth}{!}{ 
\begin{tabular}{@{}ccccc@{}}
\toprule
\multirow{3}{*}{\begin{tabular}[c]{@{}c@{}}Problem \\ Instance\end{tabular}} & \multicolumn{4}{c}{Number of non-dominated solutions} \\ \cmidrule(l){2-5} 
& \multicolumn{2}{c}{MDA} & \multicolumn{2}{c}{SB-SA} \\
\cmidrule(l){2-5} 
 & Mean & Stdev  & Mean  & Stdev  \\ \midrule
qapStr.25.0.1 & 36 & 9 & 4 & 1 \\
qapStr.25.0.2 & 27 & 10 & 4 & 1 \\
qapStr.25.0.3 & 34 & 10 & 4 & 1 \\ \midrule
qapStr.25.p75.1 & 24 & 7 & 4 & 1 \\
qapStr.25.p75.2 & 13 & 4 & 3 & 1 \\
qapStr.25.p75.3 & 13 & 4 & 2 & 1 \\ \midrule
qapStr.25.n75.1 & 44 & 8 & 5 & 1 \\
qapStr.25.n75.2 & 31 & 10 & 4 & 2 \\
qapStr.25.n75.3 & 22 & 6 & 4 & 2 \\ \midrule
qapStr.50.0.1 & 81 & 20 & 5 & 1 \\
qapStr.50.0.2 & 66 & 13 & 6 & 2 \\
qapStr.50.0.3 & 79 & 16 & 6 & 2 \\ \midrule
qapStr.50.p75.1 & 70 & 16 & 3 & 1 \\
qapStr.50.p75.2 & 54 & 15 & 4 & 1 \\
qapStr.50.p75.3 & 57 & 21 & 5 & 2 \\ \midrule
qapStr.50.n75.1 & 98 & 25 & 5 & 1 \\
qapStr.50.n75.2 & 87 & 21 & 5 & 1 \\
qapStr.50.n75.3 & 92 & 20 & 5 & 1 \\ \bottomrule
\end{tabular}}
\end{table}

\begin{table}[b]
\caption{Computation Times in Seconds\label{tb:tts}}
\centering
\resizebox{0.9\columnwidth}{!}{ 
\begin{tabular}{@{}ccccc@{}}
\toprule
\multirow{3}{*}{\begin{tabular}[c]{@{}c@{}}Problem \\ Instance\end{tabular}} & \multicolumn{4}{c}{Computation times (seconds)} \\ \cmidrule(l){2-5}
 & \multicolumn{2}{c}{MDA} & \multicolumn{2}{c}{SB-SA} \\ \cmidrule(l){2-5} 
 & Mean Time & Stdev Time & Mean Time & Stdev Time \\ \midrule
qapStr.25.0.1 & 8.38 & 0.77 & 47.26 & 0.34 \\
qapStr.25.0.2 & 8.18 & 0.66 & 45.70 & 0.19 \\
qapStr.25.0.3 & 8.37 & 0.58 & 46.47 & 0.21 \\ \midrule
qapStr.25.p75.1 & 8.30 & 0.67 & 47.38 & 0.36 \\
qapStr.25.p75.2 & 9.11 & 0.64 & 47.22 & 0.27 \\
qapStr.25.p75.3 & 8.64 & 0.67 & 47.08 & 0.28 \\ \midrule
qapStr.25.n75.1 & 8.01 & 0.63 & 51.13 & 0.55 \\
qapStr.25.n75.2 & 8.33 & 0.63 & 48.34 & 1.54 \\
qapStr.25.n75.3 & 8.39 & 0.75 & 46.30 & 0.13 \\ \midrule
qapStr.50.0.1 & 426.68 & 13.60 & 1,967.32 & 17.11 \\
qapStr.50.0.2 & 432.00 & 7.81 & 1,977.68 & 19.39 \\
qapStr.50.0.3 & 460.10 & 13.77 & 2,016.94 & 16.89 \\ \midrule
qapStr.50.p75.1 & 458.26 & 16.65 & 1,975.88 & 17.51 \\
qapStr.50.p75.2 & 343.29 & 3.51 & 2,000.59 & 20.26 \\
qapStr.50.p75.3 & 349.79 & 3.29 & 2,009.54 & 21.06 \\ \midrule
qapStr.50.n75.1 & 444.87 & 13.22 & 1,962.74 & 19.57 \\
qapStr.50.n75.2 & 455.81 & 12.03 & 2,013.71 & 26.36 \\
qapStr.50.n75.3 & 407.66 & 9.23 & 1,977.70 & 16.79 \\ \bottomrule
\end{tabular}}
\end{table}